# Performance Evaluation of Swin Vision Transformer Model using Gradient Accumulation Optimization Technique

Sanad Aburass [1] and Osama Dorgham [2,3]

[1] *Department of Computer Science, Maharishi International University, Fairfield, Iowa, USA.*
*Saburass@miu.edu*
[2] *Prince Abdullah bin Ghazi Faculty of Information and Communication Technology, Al-Balqa Applied University, 19117, Al-Salt, Jordan. o.dorgham@bau.edu.jo*
[3] *School of Information Technology, Skyline University College, University City of Sharjah – P.O. Box 1797 - Sharjah, United Arab Emirates. o.dorgham@skylineuniversity.ac.ae*

**Abstract.** Vision Transformers (ViTs) have emerged as a promising approach for visual recognition tasks, revolutionizing the field by leveraging the power of transformer-based architectures. Among the various ViT models, Swin Transformers have gained considerable attention due to their hierarchical design and ability to capture both local and global visual features effectively. This paper evaluates the performance of Swin ViT model using gradient accumulation optimization (GAO) technique. We investigate the impact of gradient accumulation optimization technique on the model's accuracy and training time. Our experiments show that applying the GAO technique leads to a significant decrease in the accuracy of the Swin ViT model, compared to the standard Swin Transformer model. Moreover, we detect a significant increase in the training time of the Swin ViT model when GAO model is applied. These findings suggest that applying the GAO technique may not be suitable for the Swin ViT model, and concern should be undertaken when using GAO technique for other transformer-based models.

## 1 Introduction

Image classification is a fundamental task in computer vision, which involves assigning a label to an image based on its content. This task has many practical applications, such as object recognition, facial recognition, and medical imaging [1], [2]. In recent years, deep learning methods, especially CNNs, have achieved remarkable success in image classification [3]–[5]. CNNs are neural networks specifically designed for processing and analyzing images, and they can capture complex patterns and features from images. CNNs consist of multiple layers, including convolutional, pooling, and fully connected layers. The convolutional layer is the core component of the CNN, which extracts features from the input image by applying a set of filters to the image. The pooling layer is used to down-sample the feature maps, reducing the spatial size of the output. The fully connected layer is used to classify the image based on the extracted features [6], [7]. Despite the success of CNNs in image classification, they have some limitations. CNNs are imperfect for modeling long-range dependencies in images [8], [9], which are crucial for understanding the context and relationships between different objects in an image. Transformers, on the other hand, are attention-based models that excel at capturing long-range dependencies in sequences, such as natural language processing [10]. Transformers have also shown promising results in image classification, especially for large datasets such as ImageNet [11]. The recent ViT model applies the Transformer architecture to image classification. ViT replaces the CNN's convolutional layers with a set of self-attention layers, which allow the model to attend to all the image pixels simultaneously, capturing the global context of the image. The Swin ViT is a recent improvement to the ViT model that addresses the limitation of long-range dependencies by using a hierarchical architecture. Swin divides the image into non-overlapping patches, which are processed by a series of self-attention layers. The resulting features are then aggregated using a Swin block, which captures both local and global dependencies [12]–[14]. Gradient Accumulation Optimization (GAO) is a technique that can be used to improve training efficiency in deep learning models. It involves accumulating the gradients over multiple mini-batches before updating the weights. This technique helps to reduce memory usage and allows for larger batch sizes, leading to faster convergence. However, its effectiveness depends on several factors like the number of the mini-batches and the learning rate, and it may not always lead to better results [15]. We have implemented the gradient accumulation optimization on the Swin ViT model and conducted experiments to measure its performance on image classification tasks using the CIFAR10 [16] and MNIST [17] datasets. Experiments involve training the Swin ViT model with and without gradient accumulation and comparing their accuracy and training time performance.

Our contributions summarized as follows: we present the possibility of applying GAO technique for classification model as in Swin ViT. We evaluate the performance (i.e. accuracy and time) of Swin Vision Transformer (ViT) model using gradient accumulation optimization (GAO) technique. To the best of our knowledge, this paper is the first to provide realistic performance evaluation of swin ViT model using such an optimization technique. The content of the paper can be summarized as follows. Section 2 presents the methodology and describes the implementation of our work. Section 3 presents the results and model evaluation. Finally, Section 4 presents the conclusions of this paper.

## 2 Methodology

### 2.1 *Data Acquisition* References

The CIFAR10 and MNIST datasets are two popular datasets used in the field of machine learning and computer vision for classification tasks. The CIFAR10 dataset consists of 60,000 32x32 color images, with 10 different classes, each containing 6,000 samples. These classes include objects such as airplanes, automobiles, birds, cats, dogs, and more. On the other hand, the MNIST dataset consists of 70,000 28x28 grayscale images of handwritten digits, with 10 different classes, each containing 7,000 samples. The classes in this dataset represent digits ranging from 0 to 9. Both datasets have been widely used in research for classification tasks, with many models achieving high levels of accuracy on these datasets.

### 2.2 *Vision Transformers*

ViT and Swin Transformer are two popular models for image classification tasks. Both models consist of two main components: the Transformer encoder and the MLP head [18]. ViT Model: The ViT model takes an image as input and transforms it into a sequence of fixed-length vectors. The Transformer encoder is composed of L layers and consists of four main steps. Firstly, the image is split into a sequence of non-overlapping patches:

$$x_i = w_{patch} \times patch_i \tag{1}$$

where $x_i$ is the $i^{th}$ patch, $patch_i$ is the representation of the $i^{th}$ patch, $w_{patch}$ is a learnable weight matrix.

Secondly, learnable position embeddings are added to each patch to encode the spatial information of the image:

$$x_i = x_i \times pos_i \tag{2}$$

where $pos_i$ is the learnable position embedding for patch $i$.

Thirdly, multi-head self-attention mechanism and feedforward neural networks are applied to the input embeddings:

$$x_{i'} = MultiHeadAtt(x_i) \times x_{i''} = FFN(x_{i'}) \times x_{i'''} = LayerNorm(x_{i'} + x_{i''}) \tag{3}$$

where $MultiHeadAtt$ is the multi-head self-attention mechanism, $FFN$ is the feedforward neural network, and $LayerNorm$ is the layer normalization function.

Lastly, the output embeddings are aggregated by taking the mean or max pooling over the sequence dimension:

$$Z = Pooling(x_{1'''}, x_{2'''}, x_{3'''}, \ldots \ldots x_{N'''}) \tag{4}$$

where $N$ is the number of patches and $Pooling$ is the mean or max pooling operation. The MLP head takes the output of the Transformer encoder as input and performs linear projection, activation, dropout, and linear projection to obtain the final classification result. The Swin Transformer model takes an image as input and transforms it into a sequence of fixed-length vectors. The Swin Transformer encoder is composed of K groups, and each group contains a set of non-overlapping patches. The patches in each group are processed by multi-layer Shifted Windows to generate a set of Swin Transformer blocks. Each Swin Transformer block consists of a Shifted Window Attention (SWA) layer, a local window-based feedforward network (LWFFN) layer, and a residual connection [19]. The output of each block is passed as input to the next block within the same group, and the output of the last block in each group is passed as input to the first block in the next group. The output embeddings are aggregated by taking the mean or max pooling over the sequence dimension:

$$Z = Pooling(x_1^L, x_2^L, x_3^L, \ldots \ldots x_N^L) \tag{5}$$

where $L$ is the number of Swin Transformer blocks in each group, and $x_{i^l}$ is the output of the $l^{th}$ block in group $K$. Then, the Swin Transformer head performs linear projection, activation, dropout, and linear projection to obtain the final classification result.

In summary, both ViT and Swin ViT models use a Transformer encoder to transform images into fixed-length vectors and an MLP head for classification. The ViT model uses a multi-head self-attention mechanism and feedforward neural networks, while the Swin ViT model uses multi-layer Shifted Windows to generate a set of Swin Transformer blocks. Both models can be trained using backpropagation with stochastic gradient descent (SGD) or other optimization methods. Figures 1 and 2 show the architectures of ViT and Swin-ViT respectively.

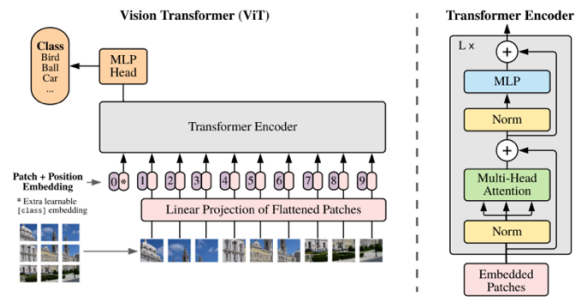

**Fig. 1.** The architecture of Vision Transformers [18]

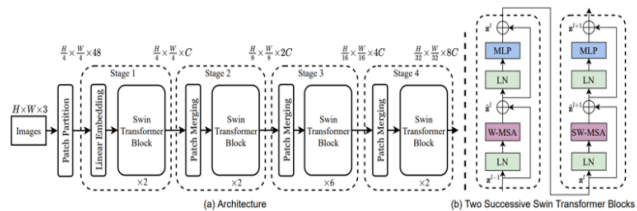

**Fig. 2.** The architecture of Swin Transformers [19]

### 2.3 *Gradient Accumulation Optimization*

Gradient accumulation optimization, also known as gradient accumulation over multiple small batches, is a technique used in deep learning to overcome the limitations of GPU memory while training deep neural networks. This technique allows the model to accumulate gradients over multiple small batches before updating the model's parameters. In this way, memory usage during training is reduced, while the model's accuracy is improved [15].

The basic idea behind gradient accumulation optimization is to perform multiple forward and backward passes on small batches of data before updating the model's parameters. Suppose we have a batch size of B, and we want to accumulate gradients over N batches. In that case, we split the original batch into N smaller batches of size B/N and perform forward and backward passes on each of these smaller batches. The gradients obtained from each backward pass are then accumulated over the N batches before updating the model's parameters.

Mathematically, the gradient accumulation optimization can be expressed as follows:
1. For each training step t, split the batch into N smaller batches of size B/N, and perform forward and backward passes on each of these smaller batches.
2. Accumulate the gradients obtained from each of the N backward passes:
$$\Delta\theta^{(t)} = \sum_{i=1}^{N} \Delta\theta_i^{(t)} \qquad (9)$$
where $\Delta\theta^{(t)}$ is the gradient obtained from the backward pass on the $i^{th}$ smallest batch.
3. After accumulating the gradients over N batches, update the model's parameters using the accumulated gradient:
$$\theta^{(t+1)} = \theta^{(t)} - \eta\Delta\theta^{(t)} \qquad (10)$$
where $\eta$ is the learning rate, and $\theta^{(t)}$ and $\theta^{(t+1)}$ are the model parameters before and after the update, respectively. The above equations illustrate the process of gradient accumulation optimization. This technique is especially useful for training large models with limited GPU memory.

### 2.4 *Experiment setup*

In our experimental setup, we utilized the resources provided by Google Colab, which offers a cloud-based environment for machine learning development. We leveraged the computational power of a GPU and 25 GB of RAM to train and evaluate our models efficiently. To build our models, we utilized Python, a popular programming language in the machine learning community, and TensorFlow, a widely used deep learning framework that provides high-level APIs for building and training deep neural networks. We chose TensorFlow because of its ease of use, its extensive documentation, and its ability to run on both CPUs and GPUs. Our experimental setup allowed us to run our experiments smoothly and efficiently, enabling us to focus on model development and analysis.

## 3. Results and Discussion

In this study, we applied gradient accumulation optimization on Swin ViT and compared its performance with the standard Swin ViT on the CIFAR10 and MNIST datasets. The results showed that using the optimization led to a decrease in accuracy and a significant increase in training time, as shown in figures 3 and 4, unlike the uplifting performance for applying GAO on [15]. We believe that the reason behind these results is overfitting, as the training accuracies were much higher than the testing accuracies, as shown in figures 5 and 6. Overfitting occurs when a model learns to fit the training data too closely, leading to poor generalization performance on new, unseen data. It can be caused by a variety of factors, such as model complexity, insufficient data, or inappropriate optimization strategies. In our case, we are dubious that the gradient accumulation optimization led to overfitting because it allowed the model to learn from the same data multiple times before updating the weights, which may have caused the model to become too specialized to the training set. One possible solution to this problem is to use regularization techniques to prevent overfitting. Regularization refers to a set of techniques that aim to reduce the model's variance by adding constraints or penalties to the optimization objective. For example, we could use L2 regularization to penalize large weights, or dropout to randomly remove units during training to prevent co-adaptation. Another approach is to use early stopping, where we stop training when the validation performance starts to deteriorate, to avoid overfitting.

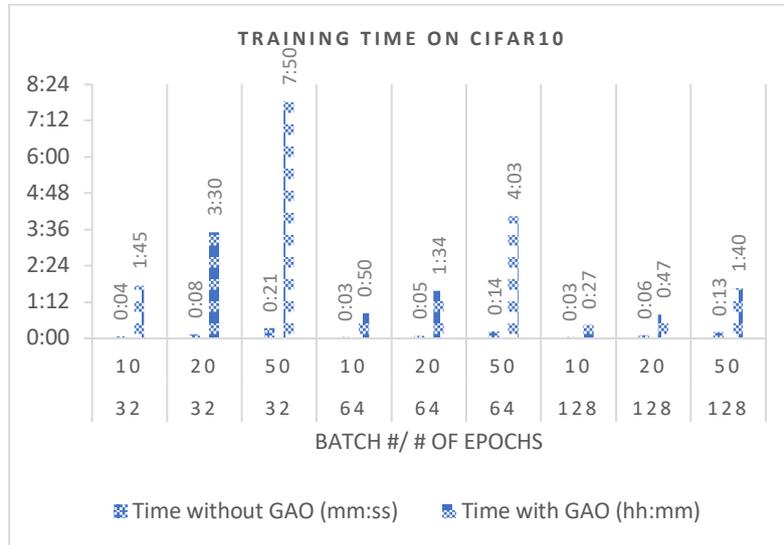

**Fig.** 3. Training Time of Swin ViT before and after applying GAO on CIFAR10

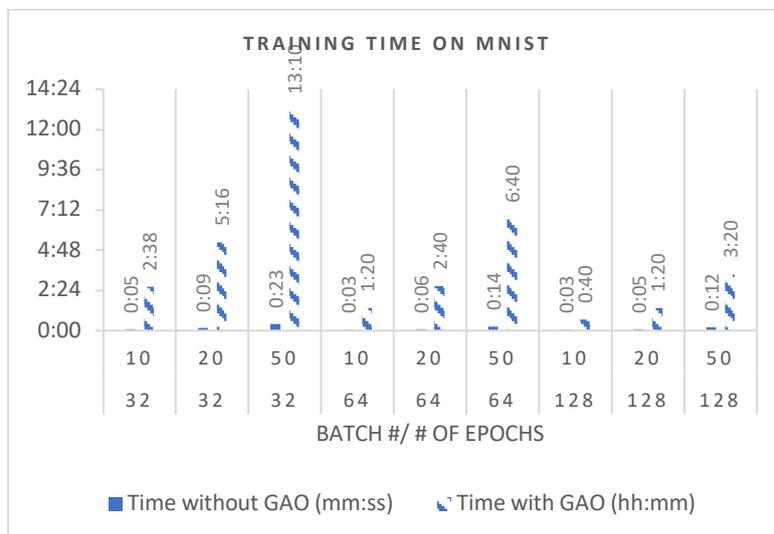

**Fig**. 4. Training Time of Swin ViT before and after applying GAO on MNIST

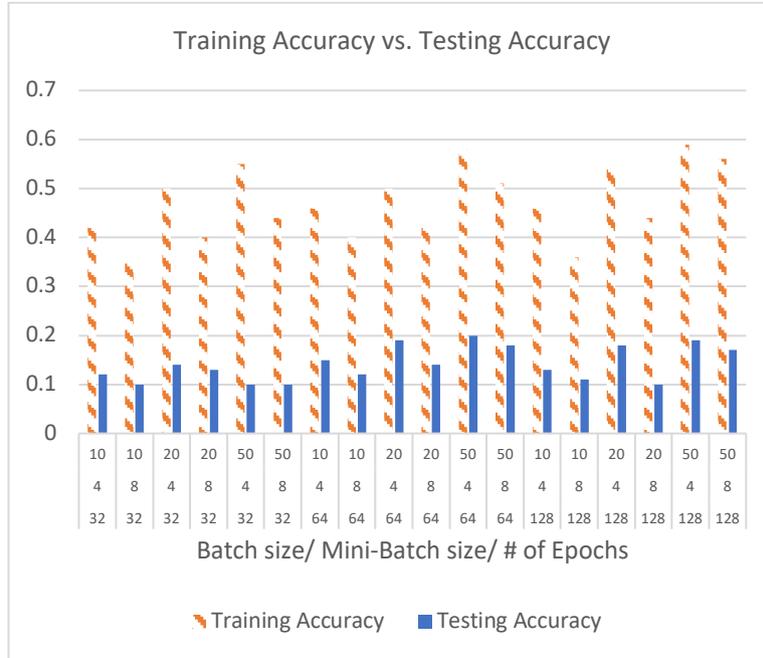

**Fig.** 5. Training accuracy and Testing accuracy of Swin ViT after applying GAO on CIFAR10

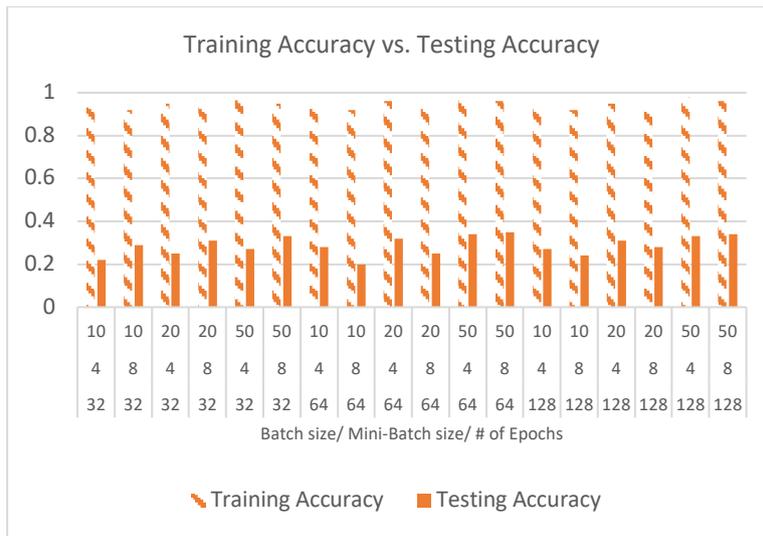

**Fig.** 6. Training accuracy and testing accuracy of Swin ViT after applying GAO on MNIST.

## 4. Conclusion

Our study evaluated the effectiveness of gradient accumulation optimization on the Swin ViT model. Our results indicate that the application of this optimization technique resulted in a considerable reduction in accuracy and significantly increased the training time compared to the standard Swin Transformers. Thus, caution should be exercised when using gradient accumulation optimization for the Swin ViT model, and other transformer-based models. Overall, our findings provide insights into the performance of gradient accumulation optimization and its potential impact on transformer-based model. Also, our study suggests that gradient accumulation optimization may not be an effective strategy for improving the performance of Swin ViT on the CIFAR10 and MNIST datasets. The observed decrease in accuracy and increase in training time may be due to overfitting caused by the optimization. Future research could explore alternative optimization strategies or regularization techniques to improve the performance of Swin ViT on these datasets.